# Predicting Rapid Fire Growth (Flashover) Using Conditional Generative Adversarial Networks

*Kyongsik Yun, Jessi Bustos, Thomas Lu, Jet Propulsion Laboratory, California Institute of Technology, Pasadena, CA*
kyun@jpl.nasa.gov

## Abstract

*A flashover occurs when a fire spreads very rapidly through crevices due to intense heat. Flashovers present one of the most frightening and challenging fire phenomena to those who regularly encounter them: firefighters. Firefighters' safety and lives often depend on their ability to predict flashovers before they occur. Typical pre-flashover fire characteristics include dark smoke, high heat, and rollover ("angel fingers") and can be quantified by color, size, and shape. Using a color video stream from a firefighter's body camera, we applied generative adversarial neural networks for image enhancement. The neural networks were trained to enhance very dark fire and smoke patterns in videos and monitor dynamic changes in smoke and fire areas. Preliminary tests with limited flashover training videos showed that we predicted a flashover as early as 55 seconds before it occurred.*

## Introduction

Reducing the number of firefighter injuries and deaths is a clear goal [1, 2]. To address this issue, the US Department of Homeland Security has launched a program called the Next Generation First Responder (NGFR) [3-5]. As a result of this program, firefighters will be fully aware of the situation with sensors and cameras that can be worn. The goal of this program is to help the first responder to be safe and healthy and to be successful in his or her mission.

Adding sensors and providing more data to the first responder can cause data overload difficulties. If the data is overwhelming, the first responder may not be able to perform important activities [6, 7]. A first responder can make a wrong decision because he/she cannot quickly extract key insights from the flood of information. This is not because there is no relevant data. Information overload is a barrier to the first responder to perform his or her job safely and efficiently.

It is important to manage the cognitive load of the first responder. Human working memory holds information for only a few seconds and holds only five to seven items at a time [8, 9]. Working memory can even manipulate information [10, 11]. Therefore, there is a great need for an automated system that can determine what important data is and how to convert it to actionable knowledge or insight. In the first responder scenario, real-time functionality is paramount and it is important to choose the information to send to the first responder. The information needs to be transformed into insight as well as relevant to the situation.

One of the most important and contextual insights for firefighters is to predict flashover phenomena [12-14]. Flashover is an example of a fire that spreads very quickly across a gap because of the intense heat. Flashover is the most frightening phenomenon among firefighters. In particular, firefighters need years of training and experience to identify, predict, and plan how to engage flashovers [15, 16].

Previous studies constructed mathematical models of flashover based on variables, including fuel, temperature, and ventilation [17-19]. A previous study applied a numerical model using oxygen concentration and mean flame surface size to predict flashover time, and the authors predicted flashover within 70 seconds before it occurred [20]. Another study developed a method to estimate the temperature of a fire before flashover [21]. However, these computational models of flashover fire require structural information of the area and temperature/oxygen data using special sensors that are not practical for real fire mission applications.

The purpose of this study is to develop a computer vision algorithm that predicts flashover fires and provides early warning to firefighters.

## Methods

The index of flashovers includes dark smoke, high heat and rollover (angel fingers) and can be quantified by color, size and shape. Firefighters use hand-held thermal imaging cameras to detect and analyze the quality of fire in dark situations [22]. Hand-held cameras require one hand for location and operation, leaving only one hand for other tasks. Due to the lack of a properly used thermal imaging camera, it has contributed to the injury and death of a firefighter [23, 24]. Moreover, due to the budget constraints of the fire department, the infrared camera system is not currently available for all firefighters [25].

To solve this problem, we used a standard body camera and analyzed the color video stream. We applied generative adversarial neural networks [26] to enhance very dark fire and smoke patterns. We then monitored the dynamic changes of smoke and fire. Finally, we provided information on when flashover could occur.

The basic idea comes from the fact that experienced firefighters can identify fire and smoke quality visually even in very dark conditions [27-29]. They can imagine a thermal image through a regular color image. So we applied image-to-image conversion techniques using conditional adversarial networks [30] that have been shown to be effective in synthesizing photos from label maps, reconstructing objects from edge maps, and colorizing images.

Generative adversarial networks learn the loss of classifying whether the output image is real or not, while at the same time the networks learn the generative model to minimize this loss [26]. The conditional adversarial networks learn the mapping function from the input image to the output image as well as learning the loss function to train the mapping [30, 31]. We also used the U-Net architecture [32] and PatchGAN [33] for the discriminator.

*Training and testing data*

Conditional adversarial network training requires a thermal image (input) that matches a regular color image (output). The network maps the input image to the output image.

Flashover training videos were provided by Lewisville Fire Department and Texas State Fire Marshal's Office. These videos show the fire generation and evolution through flashover at an early stage using the burn pod made by the department. The video includes both a regular video camera and a thermal camera view that can be seen by firefighters. A total of 40 frames of regular and thermal image pairs were used in this preliminary study (30 images for training, 10 images for the test).

*Post-processing*

We calculated the number of pixels in the red, green, and blue channels. The yellow channel is defined as the average of the red and green channels. Red is 300-500°F, yellow is 200-300°F, green is 100-200°F, and blue is 0-100°F depending on the temperature representation of the infrared camera. This drew the time variation of the number of pixels in each channel.

## Results

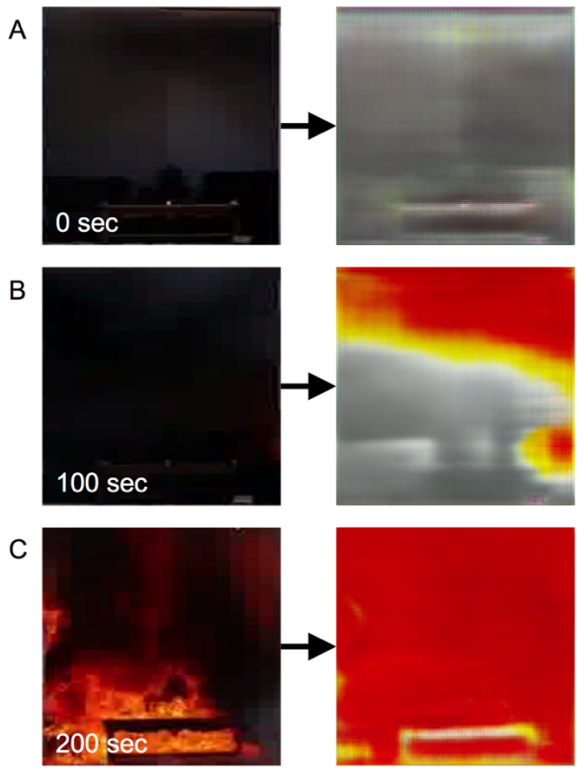

*Figure 1. Representative images of transformation from the visual to the thermal images using generative adversarial networks.*

In the test of the generative network, ten images were used as inputs from the beginning of the fire to the flashover transition (0-200 seconds). The output image was then created (Figure 1). During the early stages of the fire, the general structure of the objects (tables and sofas) and mild smoke were identified (Figure 1A). Even in very dark environments, the network successfully generated hot smoke from the input image (Figure 1B). In a flashover situation, a high heat was calculated and the table object was still identified (Figure 1C).

The output images were further analyzed by counting pixels of each thermal component (red, yellow, green, blue) and then their temporal changes were plotted (Figure 2). Based on the temporal changes of high temperature components (300-500°F and 200-300°F), we predicted a flashover as early as 55 seconds before it occurred.

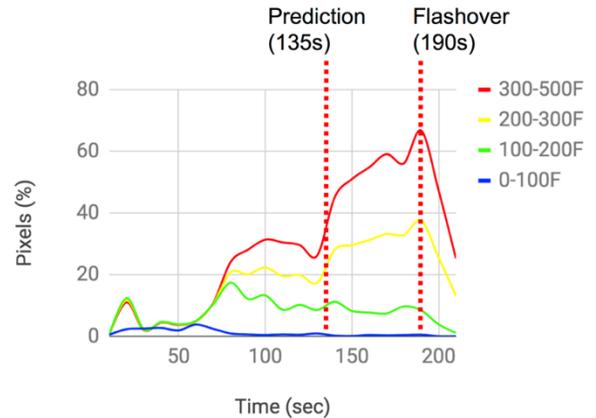

*Figure 2. Size of smoke and flame over time. Based on the rate of changes in smoke and flame size, flashover fire was predicted as early as 55 seconds before it occurred.*

## Discussion

In this study, we showed that regular body camera images can be transformed and enhanced to represent thermal information, and this technique can be used to predict flashover 55 seconds before it occurs. In order to fully understand the process of image enhancement from the regular visual image to the thermal image, we visualized the transformed intermediate results in each latent space of the networks (Figure 3). In the early part of the space, representative features (i.e., fire and smoke) were improved. In the deeper part of the space, fire and smoke were further subdivided by the estimated temperature. Then representative objects (i.e., tables and sofas) were detected. As a result, thermal information and representative objects of fire and smoke were reconstructed in the final output image. Visualization of the hidden layers of neural networks is particularly useful for understanding the internal dynamics of the whole process, which has often been considered a black box [34, 35].

Conditional adversarial networks have shown good results in small datasets (fewer than 100 samples) [30]. The advantage of a small dataset is the speed of training. The results of this study were generated based on less than 2 hours of training on a single NVIDIA GTX1080 GPU. Because the test output image transformation takes less than 100 milliseconds, this study can be applied as a real-world solution for the first responder to integrate into augmented reality glasses to provide improved insight.

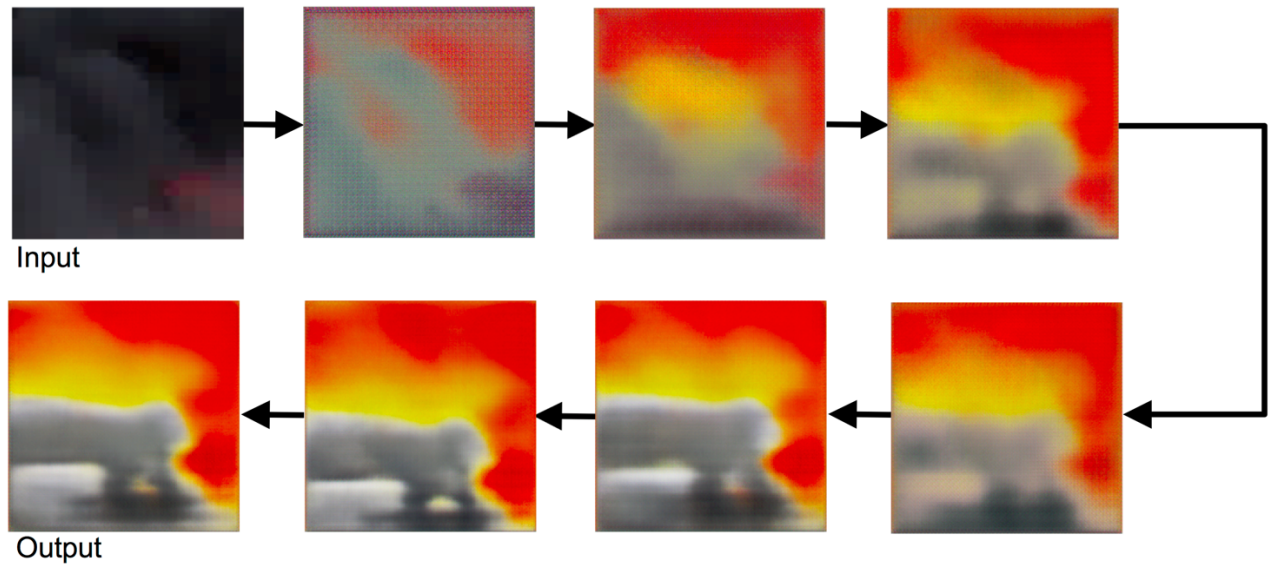

*Figure 3. Representative latent-space visualizations of the generative adversarial networks for image enhancement*

A recent study presented a thermal camera system integrated into a firefighter helmet equipped with an OLED display [36]. Our proposed algorithm can also be applied to predict flashover by automatically analyzing thermal images and quantifying the time changes of thermal content of smoke and fire. We can also use the same technique to reconstruct the occluded objects to augment the vision for firefighters. As for information overload, a high level of information (i.e., flashover timing prediction) can be provided to firefighters without overlaying the heat/occluded information on the display. This is best not to disturb the firefighter's view, but still provides the most important information.

## Acknowledgment

The research was carried out at the Jet Propulsion Laboratory, California Institute of Technology, under a contract with the National Aeronautics and Space Administration. The research was funded by the U.S. Department of Homeland Security Science and Technology Directorate Next Generation First Responders Apex Program (DHS S&T NGFR) under NASA prime contract NAS7-03001, Task Plan Number 82-106095.

## Author Biography

*Kyongsik Yun is a technologist at the Jet Propulsion Laboratory. His research focuses on building brain-inspired technologies and systems, including deep learning computer vision, natural language processing, brain-computer interfaces, and noninvasive remote neuromodulation. Kyongsik received his B.S. in Bioengineering and Ph.D. in Computational and Cognitive Neuroscience from the Korea Advanced Institute of Science and Technology (KAIST). He then worked as a postdoctoral scholar at the California Institute of Technology (Caltech).*

*Jessi Bustos is a Software Engineer at the Jet Propulsion Laboratory. His research interest includes Deep learning, A.I., computer vision, and cloud computing. Jessi received his B.S. in Physics from the University of California, Los Angeles (UCLA) and is currently in a Master's program in Computer Science at University of Southern California (USC). He is currently working under the Ground Data System Integration, Test, and Deployment group for the Mars 2020 mission.*

*Thomas Lu. is a senior researcher at the Jet Propulsion Laboratory. Thomas led the development of multi-stage neural network for automatic target recognition (ATR) and high-speed electronic interface systems for large-scale data storage projects. His research interests are in AI, deep learning, computer vision, optical sensing, and 3D imaging. Thomas received his Ph.D. degree in Electrical Engineering from the Pennsylvania State University.*